\documentclass[review,authoryear]{elsarticle}

\usepackage{amssymb}
\usepackage{graphicx}
\usepackage{hyperref}

\begin{document}

\begin{frontmatter}



\title{Conceptual Spaces for Cognitive Architectures:  \\ A \emph{Lingua Franca}
for Different Levels of Representation}
 

\author[UniTOaddress,ICARaddress,MEPhl]{Antonio Lieto}
\author[UniPAaddress,ICARaddress]{Antonio Chella}
\author[UniGEaddress]{Marcello Frixione}

\address[ICARaddress]{ICAR-CNR, Palermo, Italy}
\address[UniTOaddress]{University of Turin, Dip. di Informatica, Torino, Italy}
\address[UniPAaddress]{University of Palermo, DIID, Palermo, Italy}
\address[UniGEaddress]{University of Genoa, DAFIST, Genova, Italy}
\address[MEPhl]{National Research Nuclear University, MEPhI, Moscow, Russia}
\address[MEPhl2]{P.S. PRE-PRINT version of the paper. The final version is available at  \url{http://dx.doi.org/10.1016/j.bica.2016.10.005}}

\begin{abstract}
During the last decades, many cognitive architectures (CAs) have been realized adopting different assumptions about the organization and the representation of their knowledge level. Some of them (e.g. SOAR [\cite{laird2012soar}]) adopt a classical symbolic approach, some (e.g. LEABRA [\cite{o2000computational}]) are based on a purely connectionist model, while others (e.g. CLARION [\cite{sun2006clarion}]) adopt a hybrid approach combining connectionist and symbolic representational levels. Additionally, some attempts (e.g. biSOAR) trying to extend the representational capacities of CAs by integrating diagrammatical representations and reasoning are also available [\cite{kurup2007modeling}].
In this paper we propose a reflection on the role that Conceptual Spaces, a framework developed by Peter G\"ardenfors [\cite{gardenfors00conceptual}] more than fifteen years ago, can play in the current development of the Knowledge Level in Cognitive Systems and Architectures. In particular, we claim that Conceptual Spaces offer a \emph{lingua franca} that allows to unify and generalize many aspects of the symbolic, sub-symbolic and diagrammatic approaches (by overcoming some of their typical problems) and to integrate them on a common ground. In doing so we extend and detail some of the arguments explored by G\"ardenfors [\cite{gardenfors1997symbolic}] for defending the need of a conceptual, intermediate, representation level between the symbolic and the sub-symbolic one. In particular we focus on the advantages offered by Conceptual Spaces (with respect to symbolic and sub-symbolic approaches) in dealing with the problem of compositionality of representations based on typicality traits. Additionally, we argue that Conceptual Spaces could offer a unifying framework for interpreting many kinds of diagrammatic and analogical representations. As a consequence, their adoption could also favor the integration of diagrammatical representation and reasoning in CAs.
\end{abstract}


\end{frontmatter}


\section{Introduction}
\label{sec:Intro}

Within the field of cognitive modeling, it is nowadays widely assumed that different kinds of representation are needed in order of accounting for both biological and artificial cognitive systems. Examples are the broad class of neural network representations (including deep neural networks); the vast family of symbolic formalisms (including logic and Bayesian or probabilistic ones); analogical representations such as mental images, diagrammatic representations, mental models, and various kinds of hybrid systems combining in different ways the approaches mentioned above.

All these methods are successful in explaining and modeling certain classes of cognitive phenomena, but no one can account for all aspects of cognition. This problem also holds if we consider some recent successful artificial systems. For example, the \emph{Watson} system is based on a probabilistic system able to reason on enormous amounts of data, but it mostly fails to account for trivial common-sense reasoning (see [\cite{davis2015commonsense}], p. 94). Similarly, the \emph{AlphaGo} system [\cite{silver2016mastering}], based on massive training of deep neural networks, is impressively successful in the well-defined domain of the \emph{Go} game.  However, it is not able to transfer its approach in general or cross-domain settings. In general, this is a classical obstacle of neural networks: in order to solve a particular problem they need to be trained by a suitable and vast training set. Then, however, how to employ the learned strategies to solve similar problems is still an open issue\footnote{This issue is also explicitly reported by Hassabis in an interview published on Nature \url{http://goo.gl/9fUy4Z}.}.


Based on this evidence, our claim is that the Knowledge Level of cognitive artificial systems and architectures can take advantage of a variety of different representations. In this perspective, the problem arises of their integration in a theoretically and cognitively motivated way. While, in fact, existing hybrid systems and architectures [\cite{sun2006clarion}] are able to combine different kinds of representations (see for example the class of neuro-symbolic systems [\cite{garcez2008neural}]), nonetheless this kind of integration is usually \emph{ad hoc} based [\cite{chella1998architecture}] or, as we will show in the following sections, is only partially satisfying. Our hypothesis is that Conceptual Spaces can offer a \emph{lingua franca} that allows to unify and generalize many aspects of the representational approaches mentioned above and to integrate them on common ground. 

The paper is organized as follows: in Section 2 we report how, in Cognitive Science research, the problem of conceptual representations intended as a heterogeneous phenomenon has gained attention and experimental support in the last decades. In Section 3, we consider this pluralistic representational stance in the area of Artificial Intelligence by focusing on some of the most widely known representational approaches adopted in literature. Section 4 provides a synthetic description of Conceptual Spaces, the representational framework that we propose as a \emph{lingua franca} for the connection of the different representational levels used in different CAs. In Section 5 we outline the advantages offered by the Conceptual Spaces representation used as a grounding layer for the classical AI approaches reviewed in Section 3. In doing so we extend and detail some of the arguments explored by G\"ardenfors [\cite{gardenfors1997symbolic}] for defending the need of a conceptual, intermediate, representation level between the symbolic and the sub-symbolic one. Conclusions end the paper.

\section{Heterogeneity of Representations in Cognitive Science: The Case of Concepts}

In this Section, we present some empirical evidence from Cognitive Science that favor the hypothesis of the heterogeneity of representations in cognitive systems and architectures. In particular, we take into account two classes of evidence concerning conceptual representations: the description of non-classical concepts (Sect. 2.1) and the application of the dual process distinction to conceptual knowledge (Sect. 2.2). 

\subsection{Representing Non-Classical Concepts}
In Cognitive Science, different theories about how humans represent, organize and reason on their conceptual knowledge have been proposed.
In the traditional view, known as the classical or Aristotelian theory, concepts are defined as sets of necessary and sufficient conditions. Such theory was dominant in philosophy and psychology from the antiquity until the mid-70s of the last century, when the empirical results of Eleanor Rosch [\cite{rosch75cognitive}] demonstrated the inadequacy of such a theory for ordinary common sense concepts. These results showed that familiar concepts often exhibit \emph{typicality} effects. The results obtained by Rosch have had a crucial importance for the development of different theories of concepts trying to explain various representational and reasoning aspects concerning typicality. Usually, such theories are grouped into three broad classes: (i) the prototype theories, developed starting from the work  of Rosch; (ii) exemplars theories; and (iii) theory-theories (see e.g.  [\cite{murphy2002big}] and [\cite{machery2009doing}] for a detailed review of such approaches). All of them are assumed to account for some aspects of the typicality effects in conceptualization (such as that one of common-sense categorization).

According to the prototype view, knowledge about categories is stored using prototypes, i.e., representations of the best instance of a category. For example, the concept CAT coincides with a representation of a typical cat. In the simpler versions of this method, prototypes are represented as (possibly weighted) lists of features.

According to the exemplar view, a category is represented as set of specific exemplars explicitly stored within memory: the mental representation of the concept CAT is thus the set of the representations of (some of) the cats encountered during lifetime.

Theory-theories approaches adopt some form of holistic point of view about concepts. According to versions of theory-theories, concepts are analogous to theoretical terms in a scientific theory. For example, the concept CAT is individuated by the role it plays in our mental theory of zoology. In other versions of the approach, concepts themselves are identified with micro-theories of some sort. For example, the concept CAT is a mentally represented micro theory about cats.

Despite such approaches have been historically considered as competitors, since they propose different models, and they have different predictions about how the humans organize and reason on
conceptual information, various works (starting from Barbara Malt [\cite{malt1989line}]) showed that they are eventually not mutually exclusive. Rather, they seem to succeed in explaining different classes of cognitive phenomena. In particular, empirical data - i.e., behavioral measures as categorization probability and reaction times - suggest that subjects use different representations to categorize. Some people employ exemplars, a few rely on prototypes, and others appeal to both exemplars and prototypes. Some representations seem to be more suitable for certain tasks, or for certain categories. Also, this distinction seems to have also neural plausibility witnessed by many empirical results (the first in this line is due to Squire and Nolton [\cite{squire1995learning}]).

Such experimental results led to the development of
the so-called \emph{heterogeneous hypothesis} about the nature of conceptual representations, according to which concepts do not constitute a unitary phenomenon. In particular, different types of conceptual representations are assumed to
exist. All such representations represent different bodies of knowledge associated with the same category. Each body of conceptual knowledge is thus manipulated by various processes involved in multiple tasks (e.g. recognition, learning, categorization). 


\subsection{Dual-Process oriented Conceptual Representations}

A further divide between different kinds of conceptual representations refers to the dual process hypothesis about reasoning and rationality. According to \emph{dual process} theories [(\cite{stanovich2000advancing}], [\cite{evans2009two}], [\cite{kahneman2011thinking}]) two different types of cognitive processes and systems exist, which have been called respectively \emph{System(s) 1} and \emph{System(s) 2}.

\emph{System 1} processes are automatic. They are phylogenetically older and shared by humans and other animal species. They are innate and control instinctive behaviors, so they do not depend on training or particular individual abilities and, in general, they are cognitively undemanding. They are associative and operate in a parallel and fast way. Moreover, \emph{System 1} processes are not consciously accessible to the subject.

\emph{System 2} processes are phylogenetically recent and are peculiar to the human species. They are conscious and cognitively penetrable (i.e. accessible to consciousness) and based on explicit rule following. As a consequence, if compared to \emph{System 1}, \emph{System 2} processes are sequential and slower, and cognitively demanding. Performances that depend on \emph{System 2} processes are usually affected by acquired skills and differences in individual capabilities \footnote{A shared assumption of the dual process hypothesis is that both systems can be composed in their turn by many sub-systems and processes.}.

The dual process approach was initially proposed to account for systematic errors in reasoning. Such errors (consider, e.g., the classical examples of the selection task or the conjunction fallacy) should be ascribed to fast, associative and automatic \emph{type 1} processes, while \emph{type 2} is responsible for the slow and cognitively demanding activity of producing answers that are correct concerning the canons of normative rationality.

In general, many aspects of the psychology of concepts have presumably to do with fast, type 1 systems and processes, while others can be plausibly ascribed to type 2. 

For example, the categorization process based on typical traits (either prototypically represented or based on exemplars or theories) is, presumably, a fast and automatic process which does not require any explicit effort and which could likely be attributed to a type 1 system. On the contrary, there are types of inference that are usually included within conceptual abilities, which are slow and cognitively demanding and which should be attributed to processes that are more likely to be ascribed to type 2. Consider the ability to make explicit high-level inferences involving conceptual knowledge, and the capacity to justify them. Or consider classification: classifying a concept amounts to individuating its more specific superconcepts and its more general subconcepts, or, in other words, to identify implicit superconcept-subconcept relations in a taxonomy. For human subjects such a process is usually slow, it requires great effort, and it is facilitated by specific training. So, according to the dual process theories, the inferential task of classifying concepts in taxonomies is \emph{prima facie} a type 2 process. It is qualitatively different from the task of categorizing items as instances of a particular class on the basis of typical traits (e.g. the task of classifying Fido as a dog because it barks, has fur and wags his tail). Therefore, it is plausible that conceptual representation in cognitive systems should be assigned to (at least) two different kinds of components responsible for different tasks. In particular, type 2 processes are involved in complex and cognitively demanding inference tasks, and fast and automatic type 1 processes are involved in categorization based on the common-sense information. A similar theoretical position is defended by Piccinini [\cite{piccinini2011two}], according to which only two kinds of concept exist: implicit and explicit; he correlates implicit and explicit concepts respectively to system 1 and system 2 processes.

More recently, it has been also argued [\cite{frixione2013dealing}] that a \emph{cognitively plausible} artificial model of conceptual representation should be based on a dual process approach and, as such, formed by different components based on various representations inspired by the previous distinction. Some available systems have been developed based on this hypothesis [\cite{lieto2015common,lieto16dual}] and integrated with available Cognitive Architectures such as ACT-R \cite{anderson2004integrated}, and CLARION [\cite{sun2006clarion}]. In such systems the type 1 processes have been demanded to the Conceptual Spaces framework (Sect. 4), while the type 2 processes have been demanded to standard symbolic representations (Sect. 3.1). 
Systems whose conceptual processing activity is based on the dual process approach have been also recently investigated in the area of computational creativity of robotic agents [\cite{augello2016artwork}].

\section{Representational Formalisms and Approaches in AI}

We claim that the plurality of heterogeneous representations observed in natural cognitive systems and exemplified in the above section is also recommended in the design of cognitive artificial systems and architectures. In AI different types of representational approaches have been proposed and can be successfully used to model various aspects of the conceptual heterogeneity described above. In the following, we shortly take into account three primary examples: symbolic representations (Sect. 3.1), neural networks representations (Sect. 3.2), and diagrammatic and analogical representations (Sect. 3.3).

\subsection{Symbolic Representations} 


Symbolic representations, which in many cases rely on some logic formalism, are usually well suited for dealing with complex reasoning tasks. Such systems are characterized by the compositionality of representations: in a compositional system of representations, we can distinguish between a set of primitive, or atomic, symbols and a set of complex symbols. Complex symbols are generated from primitive symbols through the application of suitable recursive syntactic rules: generally, a potentially infinite set of complex symbols can be generated from a finite set of primitive symbols. The meaning of complex symbols can be determined starting from the meaning of primitive symbols, using recursive semantic rules that work in parallel with syntactic composition rules.

Compositionality has been considered an irrevocable trait of human cognition: in classical cognitive science, it is often assumed that mental representations are indeed compositional. A clear and explicit formulation of this assumption was proposed by Fodor and Pylyshyn as a criticism of neural networks and connectionist systems [\cite{fodor1988connectionism}]. They claim that the compositionality of mental representations is mandatory to explain fundamental cognitive phenomena (i.e., the generative and systematic character of human cognition) and that neural networks are not compositional.  

Compositionality is also a characteristic of symbolic artificial systems, and many knowledge representation formalisms are indeed compositional. In the field of Cognitive Architectures, for example, SOAR is one of the most famous systems exploiting symbolic and compositional representations of knowledge (called chunks) and using pattern matching to select relevant knowledge elements. This system adheres strictly to the Newell and Simon's physical symbol system hypothesis [\cite{newell1976computer}] which states that symbolic processing is a necessary and sufficient condition for intelligent behavior. However, compositionality cannot be easily accommodated with some cognitive phenomena. For example, it is somewhat at odds with the representation of concepts in terms of typicality [\cite{frixione2011representing}], as we shall see in greater details in Sect. 5.1 below. This problem is not limited to the empirical analysis of natural cognitive systems; it is of main relevance also for the design of cognitive artificial systems and architectures. The clash between compositionality and typicality requirements in symbolic representations is evident the field of artificial conceptual modelling. Consider, for example, description logics and ontological languages (e.g. OWL), which are fully compositional but not able to account for typicality\footnote{In the field of logic-oriented Knowledge Representation (KR) various fuzzy and non-monotonic extensions of description logics formalisms have been designed to deal with some aspects of \emph{non-classical} concepts [\cite{giordano2013non}], [\cite{straccia2002reasoning}]. Nonetheless, various theoretical and practical problems remain unsolved and, in general, an acceptable KR framework able to provide a practically usable trade-off regarding language expressivity and complexity has been not yet achieved [\cite{frixione2012representing}]. (In particular, on fuzzy logic and typicality effects, see sect. 5.1 below.)}. 
At the same time, as mentioned, representing concepts using typicality is relevant for computational applications (and in particular for those of cognitive inspiration). More in general, Fodor and Pylyshyn are true when they claim that neural networks are not compositional. However, it is also true that, in the development of cognitive artificial systems and architectures we do not want to give up with some of the advantages offered by the neural networks. It is likely that compositionality has to do with higher order cognition and with complex, \emph{type 2}, inferential tasks while neural networks are more appropriate to model \emph{type 1} phenomena.  The problem remains of the interaction of these two classes of formalisms.

In general, in the symbolic AI tradition, an attempt to mitigate this aspect has been proposed with the Bayesian networks [\cite{nielsen2009bayesian}]. Bayesian networks can be considered as a class of symbolic representations, where the relations between concepts are weighted by their strength, calculated through statistical computations. Despite the recent successes of the Bayesian approach for the explanation of many cognitive tasks [\cite{griffiths2008bayesian}], the acceptance of explaining intelligence of both natural and artificial minds in terms of \emph{Bayesian Machines} is still far from being achieved. Many forms of common-sense knowledge in human cognition do not require Bayesian predictions about what will happen or, in general, to reason probabilistically [\cite{sloman2014can}]. In addition, also in these more sophisticated cases of symbolic representations, the problematic aspects of reconciling compositionality and typicality requirements remains, as we shall see in section 5.1, unsolved. 



\subsection{Neural Networks Representations} 


Neural networks are a class of representations employed successfully in many architectures and in many difficult tasks (see for example the \emph{AlphaGo} system mentioned above). In general, in the field of CAs, this class of representation has been widely employed to deal with the “fast” behavior of a dynamic system and for aspects mainly related to learning and perception. Neural networks are particularly well suited for classification tasks. As a consequence, they are widely adopted in many pattern recognition problems in AI: typical case studies concern the recognition of handwritten letters and numbers. 

Differently from symbolic representations, neural networks receive input data directly coming from sensory systems, as images, signals, and so on, and thus the problem of grounding representations to entities in the external world (which is notoriously arduous for symbolic systems) is in some sense alleviated. The importance of neural networks for symbol grounding has been discussed by Harnad in a seminal paper  [\cite{harnad1990symbol}]. From this point of view, the main advantage of deep neural networks, and in particular of Convolutional Neural Networks, is that they are even closer to sensory data, and therefore they need less or no preprocessing of input data (see, e.g., the recent review by [\cite{lecun2015deep}]). 

However, representations based on neural networks are problematic in many senses. For example, as already anticipated above, it is challenging to implement compositionality in neural networks  [\cite{fodor1988connectionism}], [\cite{frixione1989symbols}]. Moreover, it is unclear how to implement complex reasoning and planning tasks, which are naturally modeled by symbolic formalisms. As a consequence, the typical move is to employ some hybrid neuro-symbolic systems. This is the case, for example, of the ACT-R architecture [\cite{anderson2004integrated}], that employs a sub-symbolic activation of symbolic conceptual chunks representing the encoded knowledge. In some cases, e.g. in ACT-R, this hybrid approach successfully allows to overcome, in a cognitively grounded perspective, many problems of the sub-symbolic and symbolic representations considered in isolation. In other cases, as earlier mentioned, this integration is \emph{ad hoc} based and does not provide any explanatory model of the underlying processes integrating the representations. In any case, however, the classical well-known problem of neural networks remains their opacity: a neural network behaves as a sort of black box and specific interpretation for the operation of its units and weights is far from trivial (on this aspect, see Sect. 5.2). 

\subsection{Diagrammatic and Analogical Representations} 

In the last decades, many types of representation have been proposed, both in Cognitive Science and in AI, which share some characteristic with pictures or, more in general, with diagrams and analog representations. Consider for example the debate on mental images that affects Cognitive Science since the seventies [\cite{kosslyn2006case}].
According to supporters of mental images, some mental representations have the form of \emph{pictures in the mind}. 

There are other examples of analog representations besides mental images. Consider the notion of \emph{mental model} as proposed by Philip Johnson-Lard [\cite{johnson1983mental}], [\cite{johnson2006we}].
According to Johnson-Laird, many human cognitive performances (e.g. in the field of deductive reasoning) can be better accounted for by hypothesizing the processing of analog representations called mental models, rather than the manipulation of sentence-like representations such as logical axioms and rules. For example, according to Johnson-Laird, subjects, when performing a deductive inference,  first create and merge an analog model of the premises, and then they check the resulting model to draw a conclusion. 

Many pictorial, analog or diagrammatic models have been proposed in various fields of Cognitive Science, which take advantage of forms of representations that are picture-like, in the sense that they spatially \emph{resemble} to what they represent (see e.g. [\cite{glasgow1995diagrammatic}] and, in the field of planning, [\cite{frixione2001diagrammatic}]).




This class of representations is heterogeneous, and it is surely not majoritarian if compared to the main streams of symbolic/logic based systems and of neural networks. Moreover, they lack a general theory, and, despite their intuitive appeal, a common and well understood theoretical framework does not exist. However, in spatial domains, they present various advantages. If compared to sub-symbolic models they are much more transparent; when compared with symbolic  representations, they are often more intuitive, and they avoid the need of a complete explicit axiomatization.  As mentioned above, some attempts also exists trying to embed diagrammatical representation in CAs [\cite{kurup2007modeling}].


From the empirical point of view, none of the above surveyed families of representations alone is able to account for the whole spectrum of phenomena concerning human cognition. This suggests that, also in artificial systems, a plurality of representational approaches is needed. However, the way in which these representations interact is not clear both from an empirical point and from a computational point of view.

\section{Conceptual Spaces as a \emph{Lingua Franca}} 

Our thesis is that geometrical representations, and in particular Conceptual Spaces [\cite{gardenfors00conceptual}], constitute a common language that enables the interaction between different types of representations. On one hand, they allow overcoming some limitations of the symbolic systems (see Sect. 5.1) concerning both the common sense and the anchoring problems. On the other hand, they represent a sort of blueprint useful for designing and modelling artificial neural networks in a less opaque way. Moreover, they provide a more abstract level for the interpretation of the underlying neural mechanisms (see Sect. 5.2). Finally, thanks to their geometrical nature, they offer a unifying framework for interpreting many kinds of diagrammatic and analogical representation (see Sect. 5.3).



The theory of Conceptual Spaces provides a robust framework for the internal representations in a cognitive agent. In the last fifteen years, such framework has been employed in a vast range of AI applications spanning from visual perception [\cite{chella1997cognitive}] to robotics  [\cite{chella2003anchoring}], from question answering [\cite{lieto2015common}] to music perception [\cite{chella2015cognitive}] (see [\cite{zenker2015applications}] for a recent overview). 
According to  G\"ardenfors, Conceptual Spaces represent an intermediate level of representation between the sub-symbolic and the symbolic one. The main feature of a Conceptual Space is given by the introduction of a geometrical framework for the representation of knowledge based on the definition of a number of quality dimensions describing concepts.  In brief, a Conceptual Space is a \emph{metric} space  in which entities are characterized by  quality dimensions [\cite{gardenfors00conceptual}].  In some cases, such dimensions can be directly related to perceptual information; examples of this kind are temperature, weight, brightness, pitch. In other cases, dimensions can be more abstract in nature \footnote{In this paper we will not consider the problem of the acquisition of such representations. We just mention that there are many successful approaches recently proposed in Computational Linguistics and Distributional Semantics [\cite{pennington2014glove,mikolov2013distributed,mikolov2013efficient}] aiming at learning vectorial structures, called \emph{word embeddings}, from massive amounts of textual documents. Word embeddings represent the meaning of words as points in a high-dimensional Euclidean space, and are in this sense reminiscent of Conceptual Spaces. However, they differ from Conceptual Spaces in at least two crucial ways that limit their usefulness for applications in knowledge representation. First, word embedding models are mainly aimed at modelling word-similarity, and are not aimed at providing a geometric representation of the conceptual information (and a framework able to perform forms of common-sense reasoning based, for example, on prototypes). Moreover, the dimensions of a word embedding space are essentially meaningless since they correspond, given an initial word, to the most statistically relevant words co-occurring with it, while quality dimensions in Conceptual Spaces directly reflect salient cognitive properties of the underlying domain. In this sense the word embeddings can be seen an intermediate step between the data level and the conceptual one in language-oriented technologies.}. 

To each quality dimension is associated a geometrical (topological or metrical) structure. The central idea behind this approach is that the representation of knowledge can take advantage of the geometrical structure of the Conceptual Spaces. The dimensions of a Conceptual Space represent qualities of the environment independently from any linguistic formalism or description. In this sense, a Conceptual Space comes before any symbolic characterization of cognitive phenomena.
A point in a Conceptual Space corresponds to an epistemologically primitive entity at the considered level of analysis. For example, in the case of visual perception, a point in a Conceptual Space is obtained from the measurements of the external world performed, e.g., by a camera, through the subsequent processing of the low-level vision algorithms. 

Concepts are represented as \emph{regions} in Conceptual Spaces. An important aspect of the theory is the definition of a \emph{metric function}. Following 
G\"ardenfors, the distance between two points in a Conceptual Space, calculated according to a metric function, 
corresponds to the measure of the perceived \emph{similarity} between the entities corresponding to the points themselves. For example, instances (or exemplars) of a concept are represented as points in space, and their similarity can be calculated in a natural way in the terms of their distance according to some suitable distance measure.

A further aspect of Conceptual Space theory has to do with the role of \emph{convex sets} of points in conceptualization.
According to the previously cited work by Rosch \cite{rosch75cognitive}, the so-called \emph{natural categories} represent the most informative level of categorization in taxonomies of real world entities. They are the most differentiated from one another, and constitute the preferred level for reference. Also, they are the first to be learned by children and categorization at their level is usually faster. 

G\"{a}rdenfors proposes the \emph{Criterion P}, according to which  natural categories correspond to \emph{convex sets} in some suitable Conceptual Space. As a consequence, \emph{betweenness}
is significant for natural categories, in that for every pair of points belonging to a convex set (and therefore sharing some features), all the points  \emph{between} them belong to the same set, and they
share in their turn the same features. 

Natural categories thus correspond to convex regions. In such scenario, therefore, prototypes and typicality effects taking place at the conceptual level have a natural geometrical interpretation: prototypes correspond to the geometrical centroid of the region itself. Then, given a certain concept, a degree of centrality is associated to each point that falls within the corresponding region. This level of centrality may be interpreted as a measure of its typicality. 

Conversely, given a set of $n$ prototypes represented as points in a Conceptual Space, a tessellation of the space in $n$ convex regions can be determined in the terms of the so-called \emph{Voronoi} diagrams [\cite{voronoi}]. In sum, one of the main features of Conceptual Spaces is represented by the fact that, differently from the models situated at the sub-symbolic and symbolic level, they provide a natural way of explaining typicality effects on concepts. Their geometrical structure allows a natural way of calculating the semantic similarity among concepts and exemplars by using classical topological (e.g., based on the Region Connection Calculus [\cite{GardenforsWilliams}]) or metrical distances. 

G\"{a}rdenfors mostly concentrated on the representation of typicality concerning prototypes. However, Conceptual Spaces allow in a natural way the representation of non-classical concepts also in terms of exemplars [\cite{frixione2013dealing}] (as we said above in Sect. 2.1, prototypes and exemplars are two complementary approaches that can explain different aspects of typicality).   




\section{On the Advantages of Conceptual Spaces} 

In the following Sections, we outline some of the advantages offered by Conceptual Spaces representations in dealing with the problems posed by the representational formalisms overviewed previously. Such analysis supports our claim that a grounding of the outlined representations in terms of Conceptual Spaces could overcome some of their limitations.

\subsection{Prototypes and Compositionality in Symbolic Representations} 


As we anticipated, compositionality can be hardly accommodated with typicality effects. In this Section, we shall argue that Conceptual Spaces could allow reconciling these two important aspects of conceptual representations. According to a well-known argument ([\cite{fodor1981present}]; [\cite{osherson1981adequacy}]), prototypes are not compositional. In brief, the argument runs as follows: consider a concept like \emph{pet fish}. It results from the composition of the concept \emph{pet} and of the concept \emph{fish}. However, the prototype of \emph{pet fish} cannot result from the composition of the prototypes of a pet and a fish: a typical pet is furry and warm, a typical fish is grayish, but a typical pet fish is neither furry and warm nor grayish. 

Let us consider a version of this argument against the possibility of reconciling compositionality and typicality effects in symbolic systems that dates back to Osherson and Smith [\cite{osherson1981adequacy}]. Osherson and Smith's original aim was to show that fuzzy logic is inadequate to capture typicality, but, as we shall see, the effect of the argument is general. At first sight, fuzzy logic seems to be a promising approach to face the problem of typicality. Indeed, one consequence of typicality effects is that some members of a category $C$ turn out to be better (i.e. more typical) instances of $C$ than others. For example, a robin is a better example of the category of birds than, say, a penguin or an ostrich. More typical instances of a category are those that share a greater number of characteristic features (e.g. the ability to fly for birds, having fur for mammals, and so on). The fuzzy value of a predicate (say, $F$) could be interpreted as a measure of typicality.  In facts, given two individuals $h$ and $k$, it is natural to assume that $F(h) > F(k)$ iff $h$ is a more typical instance of $F$ than $k$.

\begin{figure}[ht]
\centering
\includegraphics[width=.5\linewidth]{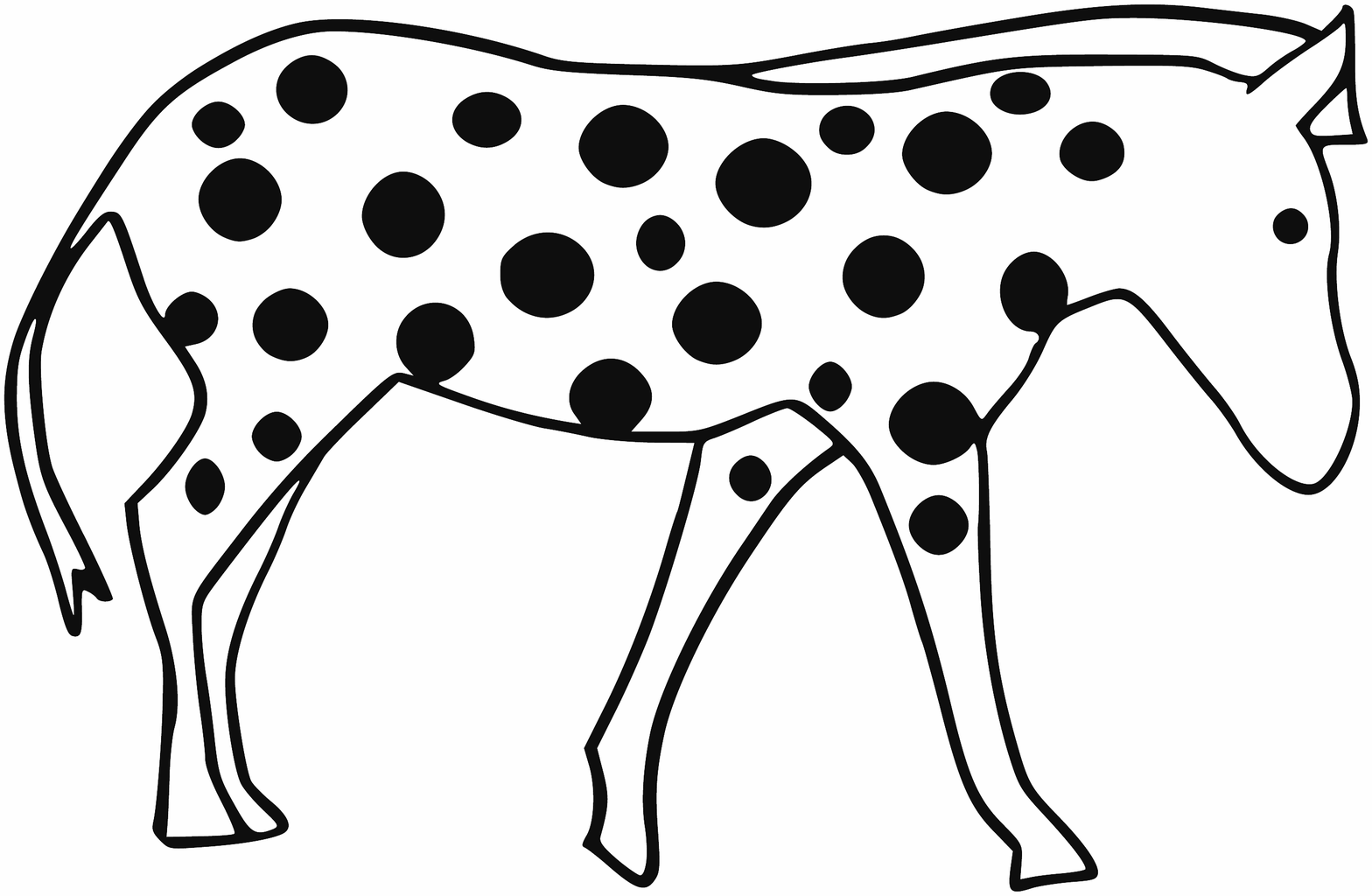}
\caption{An exemplar of the concept of \emph{Polka Dot Zebra}}
\label{fig:zebra}
\end{figure}
 
However, let us consider the zebra in Fig. 1 (and let us suppose that her name is Pina).

Pina is presumably a good instance of the concept $polka\_ dot\_ zebra$; therefore, if such a concept were represented as a fuzzy predicate, then the value of the formula $polka\_ dot\_ zebra(Pina)$ should be close to 1, say:

\begin{equation}
polka\_dot\_ zebra(Pina) = .97
\end{equation}

On the other hand, Pina is a rather poor (i.e. atypical) instance of the concept zebra; therefore the value of the formula $zebra(Pina)$ should be low, say:

\begin{equation}
zebra(Pina) = .2
\end{equation}

\noindent (of course, the specific values are not relevant here; the point is that Pina is more typical as a $polka\_dot\_ zebra$ than as a $zebra$). But $polka\_dot\_ zebra$ can be expressed as the conjunction of the concepts $zebra$ and $polka\_dot\_thing$; i.e. in logical terms, it holds that:

\begin{equation}
\forall{x} (polka\_dot\_zebra(x) \leftrightarrow zebra(x) \land  polka\_dot\_thing(x))
\end{equation}

Now, the problem is the following: if we adopt the simplest and most widespread version of fuzzy logic, then the value of a conjunction is calculated as the minimum of the values of its conjuncts. Thus,  it is impossible for the value of $zebra(Pina)$ to be .2 and that of $polka\_dot\_zebra(Pina)$ to be .97 at the same time. Of course, there are other types of fuzzy logic, in which the value of a conjunction is not the minimum of the values of the conjuncts. However, a conjunction cannot exceed the value of its conjuncts. Worse still, in general in logic, once a suitable order has been imposed on truth values, it holds that:

\begin{equation}
val(A \land B)  \leq val(A) \textrm{ and } val(A \land B) \leq val(B)
\end{equation}

So, the problem pointed out by Osherson and Smith does not seem to concern fuzzy logic only. Rather, Osherson and Smith's argument seems to show that, in general, logic-based representations are unlikely to be compatible with typicality effects\footnote{The arguments holds also if we consider to model the indicated situation in terms of a Bayesian network where the strength of the weighted value of the symbolic node polka\_dot\_zebra is assumed to be composed by the values of two nodes zebra and polka\_dot\_thing, as indicated in the example.}. Moreover, logic-based representations are paradigmatic examples of compositional systems, which fully embody the Fregean principle of compositionality of meaning. 

Indeed, the situation is more promising if, instead of logic, we face typicality by adopting a geometrical representation based on Conceptual Spaces. As previously stated, if we represent a concept as a convex area in a suitable Conceptual Space, then the degree of typicality of a certain individual can be measured as the distance of the corresponding point from the center of the area. The conjunction of two concepts is represented as the intersection of the two corresponding areas, as in Fig. 2.

\begin{figure}[ht]
\centering
\includegraphics[width=.9\linewidth]{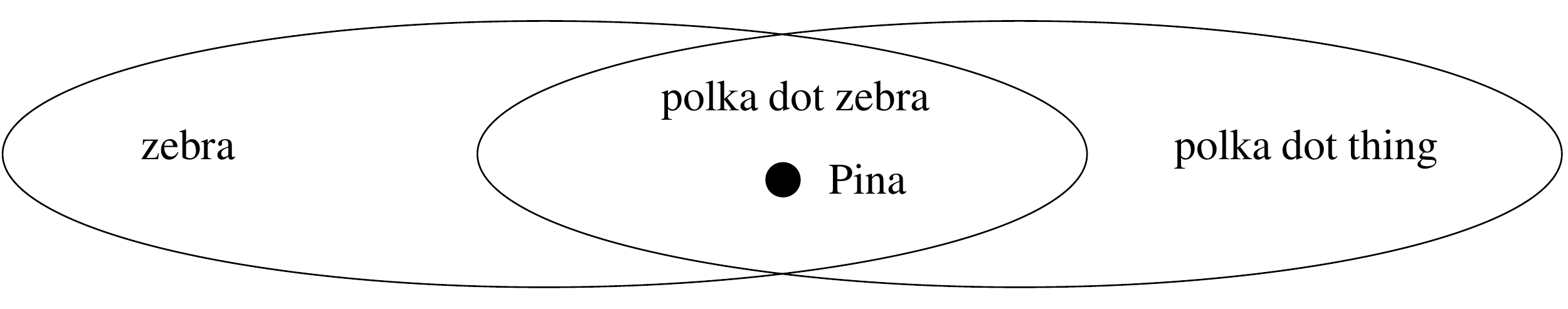}
\caption{Compositionality of Prototypes in Conceptual Spaces}
\label{fig:compositionality}
\end{figure}

According to the conceptual space approach, Pina should presumably turn out to be very close to the center of 
$polka\_dot\_zebra$ (i.e. to the intersection between $zebra$ and $polka\_dot\_thing$). In other words, she should turn out to be a very typical $polka\_dot\_zebra$, despite being very eccentric on both the concepts $zebra$ and $polka\_dot\_thing$; that is to say, she is an atypical $zebra$ and an atypical $polka\_ dot\_thing$. 
This representation better captures our intuitions about typicality. We conclude that the treatment of compositionality and that of some forms of typicality require rather different approaches and forms of representation, and should therefore presumably be assigned to different knowledge components of a cognitive architecture.  


\subsection{Interpretation of Neural Networks} 

As mentioned, neural networks, although successful in many difficult tasks are particularly well suited for classification tasks and have been widely adopted in CAs. One of the well-known problems of this class of representations is their opacity. A neural network behaves as a sort of black box: specific interpretation is troublesome for the operations of units and weights. In many cases, this is arduous to accept. Let us consider for example the case of medical domain, where it is not sufficient to classify the symptoms of a disease but it is also required to provide a detailed explanation for the provided classification of symptoms. This problem is much more pressing if we consider the case of deep neural networks where, because of the huge number of the hidden layers, there are much more units and weights to interpret. The opacity of this class of representations is also unacceptable in CAs aiming at providing transparent models of human cognition and that, as such, should be able not only to \emph{predict} the behavior of a cognitive artificial agent but also to \emph{explain} it.

A possible way out to this problem is represented by the interpretation of the network in terms of a more abstract geometric point of view. It is true that it is feasible to have a  simple geometric interpretation of the operation of a neural network: in fact, the operation of each layer may be described as a functional geometric space where the dimensions are related to the transfer functions of the units of the layer itself. In this interpretation, the connection weights between layers may be described in terms of transformation matrices from one space to another. 

However, while the interpretation of the input and output spaces depends on the given training set and the particular design of the network, the interpretation of the hidden spaces is typically tough. However, the literature reports sparse cases where a partial interpretation of the operations of the units is possible: a recent example is reported by  Zhou  [\cite{zhou2014object}]. A more general attempt to interpret the activity of a neural network in terms of information geometry is due to  [\cite{amari2007methods}].

We claim that the theory of Conceptual Spaces can be considered as a sort of designing style that helps to model more transparent neural networks, and it can facilitate the grounding and the interpretation of the hidden layers of units. As a consequence, the interpretation of neural network representations in terms of Conceptual Spaces provides a more abstract and transparent view on the underlying behavior of the networks.

G\"ardenfors [\cite{gardenfors00conceptual}] offers a simple analysis of the relationship between Conceptual Spaces and Self Organising Maps. Hereafter, Balkenius [\cite{balkenius1999there}] proposes a more articulate interpretation of the widely adopted RBF networks in terms of dimensions of a suitable Conceptual Space. According to this approach, a neural network built by a set of RBF units can be interpreted as a simple Conceptual Space described by a set of integral quality dimensions. Consequently, a neural network built by a set of sets of RBF units may be geometrically interpreted by a conceptual space made up by sets of integral dimensions.

Additionally, following the Chorus of Prototypes approach proposed by Edelman [\cite{edelman1995representation}], the units of an RBF network can be interpreted as prototypes in a suitable Conceptual Space. This interpretation enables the measurement of similarity between the input of the network and the prototypes corresponding to the units. Such an interpretation would have been much more problematic by considering the neural network alone, since this information would have been implicit and hidden. Moreover, it is possible to take into account the delicate cases of Chimeric entities, which are almost equidistant between two or more prototypes. For example, a Chimera is a lion with a goat head, and therefore, it results equidistant between the prototype of the lion and goat prototypes (see  [\cite{edelman1995representation}]). This aspect is related to the example of the polka dotted zebra provided in the previous Section. In this respect, the capability of accounting for the compositionally based on typicality traits seems to be a crucial feature of the Conceptual Spaces empowering both symbolic and sub-symbolic representations \footnote{It is worth-noting that also some forms of neuro-symbolic integration currently developed in CAs like ACT-R, and belonging to the class of the \emph{neo-connectionist} approaches, allows to deal with the the above mentioned problem by providing a series of mechanisms that are able to deal with limited forms of compositionality in neural networks  [\cite{o2013limited}] and that can be integrated with additional processes allowing the compatibility with typicality effects. In this respect, such approaches play an equivalent role to that one played by the Conceptual Spaces on these issues. In addition, however, we claim that Conceptual Spaces can offer a unifying framework for interpreting many kinds of diagrammatic and analogical representations (see section 5.3). On these classes of representations, limited work has been done by these hybrid neuro-symbolic systems (including ACT-R) [\cite{matessa2007using}]. This is a symptom that the treatment of their representational and reasoning mechanisms is not trivial in these environments and that often they need to be integrated with external diagrammatic representation systems [\cite{matessa2007using}].}.   

Finally, a recent work going in the direction of a more abstract interpretation of neural representations, and describing how the population of neural representations can be interpreted as representing vectors obtained through different kind of operations (e.g. compression and recursive binding by using circular convolutions, see  [\cite{crawford2015biologically}],) is obtained by the Semantic Pointers Perspective adopted by the NEF (Neural Engineering framework) [\cite{eliasmith2004neural}] and representing the core of the biologically inspired SPAUN architecture [\cite{eliasmith2012large}]. Such perspective is completely compatible with our proposal of providing a more abstract interpretation of neural mechanisms and representations through multidimensional Conceptual Spaces.










\subsection{Unifying Analogical and Diagrammatic Representations} 

Analogical and diagrammatic representations allow representing in an efficient and intuitive way kinds of information that would require a very complex and cumbersome amount of details if explicitly represented by symbolic and logic-oriented declarative formalisms. Let us consider a simple example that has been discussed by Philip Johnson-Laird [\cite{johnson1983mental}].
The relation \emph{to be to the right of} is usually transitive: if A is to the right of B and B is to the right of C then A is to the right of C. But consider the case in which A, B and C are arranged around, say, a small circular table. In this case, it can happen that C is to the right of B, B is to the right of A but C is not to the right of A: A and C are opposite (see Fig. 3 below).

\begin{figure}[ht]
\centering
\includegraphics[width=.5\linewidth]{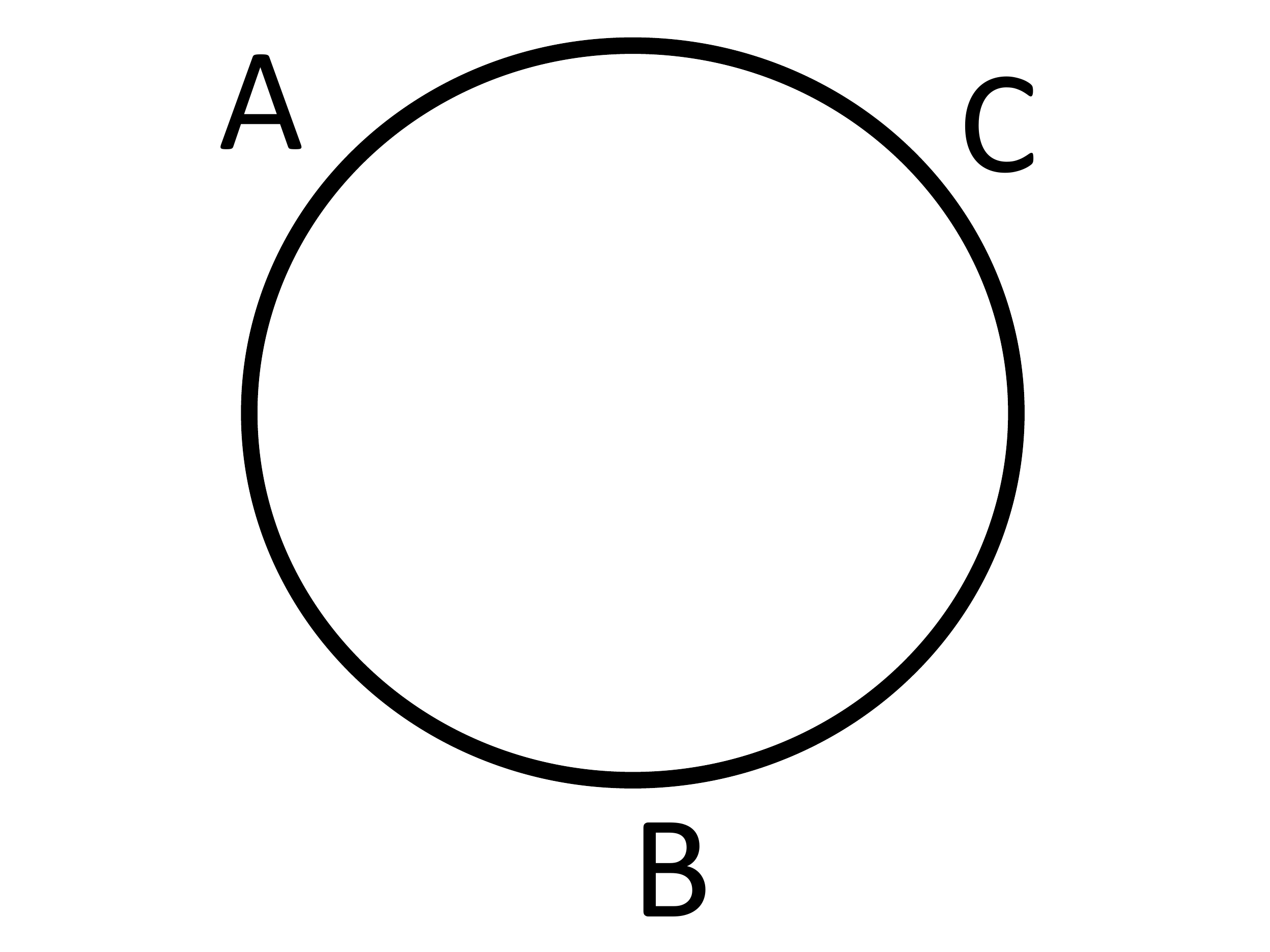}
\caption{The round table problem}
\label{fig:table}
\end{figure}

To account for such a simple fact by symbolic axioms or rules would require making explicit a huge number of detailed and complicated assertions. Conversely, the adoption of some form of analogic representation such as mental models associated with suitable procedures, e.g. procedures for the generation, revision, and inspection of mental models, would allow facing the problem in a more natural and straightforward way. 

As previously said in Sect. 3.3, a plethora of different kinds of diagrammatic representations has been proposed without the development of a unifying theoretical framework. Conceptual Spaces, thanks to their geometrical nature, allow the representation of this sort of information and offer, at the same time a general, well understood and theoretically grounded framework that could enable to encompass most of the existing diagrammatic representations. 

The geometrical nature of conceptual spaces can be useful also in representing more abstract and non-specifically spatial domains and phenomena. A typical problem of both symbolic and neural representations regards the ability to track the identity of individual entities over time. The properties of an entity change across the time. At which condition can we re-identify an entity as the same, despite its changes? In many cases the answer is not easy. Conceptual Spaces suggest a way to face the problem. We said that individual objects are represented by points in Conceptual Spaces. However, in a dynamic
perspective, objects can be rather seen as trajectories in a suitable Conceptual Space indexed by time, since the properties
of objects usually change with time. Objects may move, may age, an
object can alter its shape or color, and so on. As the properties
of an object are modified, the point, representing it in
the Conceptual Space, moves according to a certain trajectory. Since usually this modifications happens smoothly and not abruptly, 
several assumptions can be made on this trajectory,
e.g., smoothness, and obedience to physical laws [\cite{chella2004perceptual}]. 

Figuring out the evolution of an object as its future position,
or the way in which its features are going to change, can
be seen as the extrapolation of a trajectory in a Conceptual
Space. To identify again an object that has been occluded
for a certain time interval amounts to interpolate its past and
present trajectories. In general, this characteristic represents a powerful heuristic to track the identity of an individual object. Also in this case, crucial aspects of diagrammatic representations find a more general and unifying interpretation regarding Conceptual Spaces.
 






\section{Conclusions}
We have proposed Conceptual Spaces as a sort of \emph{lingua franca} allowing to unify and integrate on a common ground the symbolic, sub-symbolic and diagrammatic approaches and to overcome some well-known problems specific to such representations. In particular, by extending and detailing some of the arguments proposed by G\"ardenfors [\cite{gardenfors1997symbolic}] for defending the need of a conceptual, intermediate, representation level between the symbolic and the sub-symbolic one we have shown how Conceptual Space allow dealing with conceptual typicality effects, which is a classic problematic aspect for symbolic and logic-oriented symbolic approaches. Moreover, Conceptual Spaces enable a more transparent interpretation of underlying neural network representations, by limiting the opacity problems of this class of formalism, and it may constitute a sort of blueprint for the design of such networks. Finally, Conceptual Spaces offer a unifying framework for interpreting many kinds of diagrammatic and analogical representation. 

Our proposal may be of particular interest if we consider designing a general cognitive architecture where all these types or representation co-exist, as it is also assumed in the current experimental research in Cognitive Science \footnote{In [\cite{chella2012general}] it is also discussed the possible role that such architectural perspective can play to perform meta-computation, another crucial element for cognitive architectures.}. In this case, the Conceptual Spaces offer the common ground where all the representations find a theoretically and geometrically well-founded interpretation. 

\section{Acknowledgements}

We thank Salvatore Gaglio and Peter G\"ardenfors for the discussions on the topics presented in this article.

\section{References}



\bibliographystyle{elsarticle-harv} 


\bibliography{csr_new}


%
%
%
\end{document}